\PassOptionsToPackage{numbers,sort&compress}{natbib}
\documentclass{article}

\usepackage[preprint]{neurips_2026}
\usepackage[numbers]{natbib}

\usepackage[utf8]{inputenc}
\usepackage[T1]{fontenc}
\usepackage{hyperref}
\usepackage{url}
\usepackage{booktabs}
\usepackage{amsfonts}
\usepackage{amsmath}
\usepackage{nicefrac}
\usepackage{microtype}
\usepackage{enumitem}
\usepackage{xcolor}
\usepackage{graphicx}
\usepackage{multirow}
\usepackage{array}
\usepackage{subcaption}
\usepackage{makecell}

\title{XWOD: A Real-World Benchmark for Object Detection under Extreme Weather Conditions}

\author{%
  Chih-Hsin Chen, Yu-Tung Liu, Amar Fadillah, Kuan-Ting Lai \\
  Department of Electronic Engineering\\
  National Taipei University of Technology \\
  \texttt{\{t111C71102, t113368501, t112999405, ktlai\}@ntut.edu.tw} \\
  \And 
  Dong Liu \\
  Adobe Inc. \\
  \texttt{dongliu.hit@gmail.com}
}

\begin{document}

\maketitle

\begin{abstract}
Autonomous driving and intelligent transportation systems remain vulnerable under extreme weather. The U.S. Federal Highway Administration reports that roughly 745,000 crashes and 3,800 fatalities per year are weather-related, and recent regulatory investigations have examined failures of Level-2/3 driving systems under reduced-visibility conditions. However, datasets commonly used to evaluate weather robustness remain limited in scale, diversity, and realism. In this paper, we introduce XWOD (Extreme Weather Object Detection), a large-scale real-world traffic-object detection benchmark containing 10,010 images and 42,924 bounding boxes across seven extreme weather conditions: rain, snow, fog, haze/sand/dust, flooding, tornado, and wildfire. The dataset covers six traffic-object categories, including car, person, truck, motorcycle, bicycle, and bus. XWOD extends the weather taxonomy from one to seven conditions, and is the first to cover the emerging class of climate-amplified hazards, such as flooding, tornado, and wildfire. To evaluate the quality of our data, we train standard YOLO-family detectors on XWOD and test them zero-shot on external weather benchmarks, achieving mAP$_{50}$ scores of 63.00\% on RTTS, 59.94\% on DAWN, and 61.12\% on WEDGE, compared with the corresponding published YOLO-based baselines of 40.37\%, 32.75\%, and 45.41\%, respectively, representing relative improvements of 56\%, 83\%, and 35\%. These cross-dataset results show that XWOD provides a strong source domain for learning weather-robust traffic perception. We release the dataset, splits, baseline weights, and reproducible evaluation code under a research-use license. 
\end{abstract}

\section{Introduction}

%\paragraph{Why weather, why now.}
In recent years, camera-based perception systems have been widely deployed in passenger vehicles, autonomous taxis, and delivery robots, but their reliability under extreme weather remains limited \cite{Zang2019ImpactWeather}. This gap is safety-critical: the  five-year average data of U.S. Federal Highway Administration (FHWA) 2019--2023 attribute about 12\% of crashes, 9\% of traffic fatalities, and 11\% of injuries to weather-related events \citep{fhwa}. In October 2024, the NHTSA Office of Defects Investigation opened a preliminary evaluation covering approximately 2.4 million Tesla vehicles after four Full Self-Driving crashes under reduced roadway visibility, including one pedestrian fatality \citep{nhtsa_pe24031}. Similar failures have also been observed in commercial robotaxi deployments in dense fog \citep{waymo_sf_fog_2023}. Recent robotaxi deployments further indicate that extreme weather remains a practical operational challenge. Waymo has temporarily paused or adjusted services during flash-flood conditions in San Francisco and San Antonio, and reports from Austin suggest that standing water can still affect robotaxi operation on public roads \citep{waymo_sf_flood_2025,waymo_sa_flood_2026,waymo_sa_flood_austin_2026}. These cases show that adverse-weather perception remains a major robustness bottleneck for real-world autonomy. Moreover, climate change is broadening the operational distribution that perception systems must handle, including wildfire smoke, flash flooding, urban inundation, and severe convective weather events \citep{noaa_bdd}.

%\paragraph{What existing benchmarks miss.}
Existing extreme-weather object detection benchmarks remain limited in four aspects, as summarized in Table~\ref{tab:dataset_scale}. First, the scale is small: real-image datasets range from 1{,}000 images in DAWN \citep{dawn2020} to 4{,}322 images in RTTS \citep{reside2019}. Second, the weather coverage is narrow. Most datasets focus on a single condition, such as haze in RTTS or fog in MAS, while multi-weather datasets such as ACDC and DAWN cover only four conditions. They do not include emerging climate-amplified hazards such as flooding, tornadoes, and wildfires. Third, existing datasets have strong geographic bias: RTTS is collected in China, MAS \citep{sakaridis_eccv2018} in Zurich, and ACDC \citep{acdc2021} mainly in Central Europe, which limits their ability to evaluate cross-region generalization to North American and Asian traffic environments. Fourth, data provenance is often constrained. Synthetic datasets such as WEDGE reduce collection cost, but they may inherit the biases of their generative pipelines, and prior work has shown that models trained only on synthetic data transfer imperfectly to real adverse-weather images \citep{sakaridis_ijcv2018}.

\begin{figure}[t]

\includegraphics[width=1\linewidth]{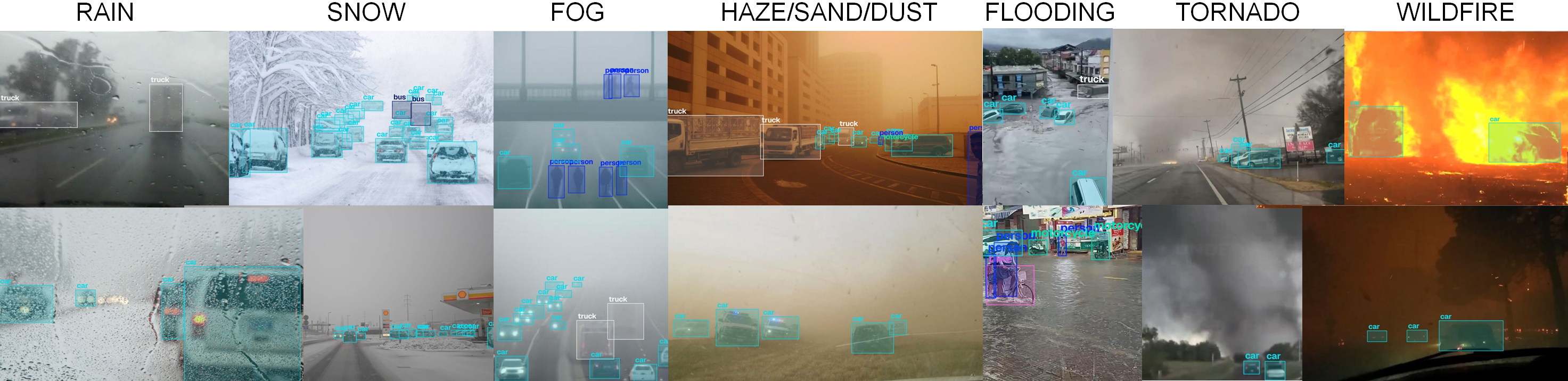}
\caption{Representative samples of XWOD: The XWOD dataset provides a wide range of real-world extreme weather images with precise annotations, capturing rare, high-impact events such as floods, tornadoes, and wildfires to push robust computer vision boundaries.}
\label{fig:examples}
\end{figure}

In order to address these limitations, we created a new dataset called XWOD (eXtreme Weather Object Detection), containing 10{,}010 images and 42{,}924 object annotations. The sample images are shown in Figure \ref{fig:examples}. Our dataset is approximately $2.3\times$ larger than RTTS, $2.5\times$ larger than ACDC, and $10\times$ larger than DAWN. To our knowledge, XWOD is the first real-image object detection benchmark that includes tornado, flooding, and wildfire scenes. It is designed to evaluate object detection robustness under real-world weather distribution shifts, especially rare and climate-amplified hazards.

%\begin{figure}[h]
%\centering
%\includegraphics[width=0.7\linewidth]{ewod_total_imaes.png}
%\caption{Image counts for real-image extreme-weather traffic-detection datasets. XWOD contains 10,010 images across seven extreme weather conditions, and covering flooding, tornado, and wildfire.}
%\label{fig:scale}
%\end{figure}

The main contributions of this work are summarized as follows:
\begin{itemize}
    \item We introduce XWOD, the largest real-image traffic object detection benchmark for extreme weather, containing 10{,}010 images and 42{,}924 bounding boxes across seven weather types, six traffic categories, and four continents.

    %\item We provide \emph{XWOD-Net}, a weather-aware detection framework evaluated across the YOLOv8, YOLOv11, and YOLOv26 families with five model scales, achieving 54.69\% mAP$_{50}$ on the in-domain XWOD test set.

    \item We demonstrate strong zero-shot cross-dataset transfer. XWOD-trained detectors improve mAP$_{50}$ from 40.37\% to 63.00\% on RTTS, from 32.75\% to 59.94\% on DAWN, and from 45.41\% to 61.12\% on WEDGE, corresponding to relative improvements of 56\%, 83\%, and 35\%, respectively.

    \item We show that XWOD is a discriminative benchmark beyond dataset scaling. Performance is not determined solely by model size or architecture, since newer YOLO families and larger variants do not consistently outperform earlier or smaller models. This suggests that XWOD evaluates robustness to diverse weather-induced domain shifts rather than only measuring model scaling effects.

\end{itemize}

We release the dataset, splits, baseline weights, datasheet, and reproducible evaluation code under a research-use license on Kaggle\footnote{\url{https://www.kaggle.com/datasets/kuantinglai/exwod}}.

\section{Related Work}
\label{sec:related}

\paragraph{Extreme-weather driving datasets.}
\emph{Foggy Cityscapes} \citep{sakaridis_ijcv2018} and \emph{Foggy Zurich / MAS} \citep{sakaridis_eccv2018} established the synthetic-vs-real-fog paradigm. RTTS \citep{reside2019} provides 4{,}322 real hazy images and is still the most-used real-haze detection testbed. ACDC \citep{acdc2021} supplies 4{,}006 semantic-segmentation frames evenly divided across fog/night/rain/snow with clear-weather correspondences. DAWN \citep{dawn2020} contains 1{,}000 web-sourced images across fog, snow, rain, and sandstorms. WEDGE \citep{wedge2023} replaces collection with DALL-E generation, yielding 3{,}360 images across 16 labels, but images retain VLM-style hallucinations and copyright provenance questions. Multimodal large-scale sensor datasets include \emph{Seeing Through Fog} \citep{stf2020}, \emph{CADC} \citep{cadc2021}, \emph{Ithaca365} \citep{ithaca365}, \emph{Boreas} \citep{boreas2023}, which focus on multimodal fusion and 3D perception rather than camera-based 2D detection. Those datasets are complementary but not comparable. BDD100K \citep{bdd100k} includes weather tags but is dominated by clear-weather frames. XWOD is purely 2D camera detection and is the first dataset to cover climate-amplified hazards (flooding, tornado, wildfire).

\paragraph{Object detectors.}
Two-stage detectors \citep{fasterrcnn,cascadercnn} remain common baselines in domain-adaptation studies. DETR and descendants \citep{detr,deformable_detr,dino_detr,codetr} and their real-time variants RT-DETR \citep{rtdetr}, D-FINE \citep{dfine}, and RF-DETR \citep{rfdetr} push accuracy-latency frontiers. The YOLO family remains the most widely deployed in production driving stacks; recent releases include YOLOv7 \citep{yolov7}, YOLOv8 \citep{yolov8}, YOLOv9 \citep{yolov9}, YOLOv10 \citep{yolov10}, YOLOv11 \citep{yolo11}, YOLOv12 \citep{yolov12}, and YOLOv26 \citep{yolo26}. We benchmark YOLOv8, YOLOv11, and YOLOv26 since they bracket the current production landscape.

\paragraph{Weather-aware detection.}
Adaptive and restoration-coupled detectors explicitly train for weather correction: IA-YOLO \citep{iayolo} integrates differentiable image processing; TogetherNet \citep{togethernet} couples restoration and detection via dynamic feature enhancement; DENet \citep{denet} adds a Laplacian-pyramid enhancement front-end; DSNet \citep{dsnet} performs joint semantic learning. Domain-Adaptive Faster R-CNN \citep{dafrcnn} pioneered adversarial adaptation for object detection under foggy weather on Cityscapes.

\section{The XWOD Dataset}
\label{sec:dataset}

\subsection{Design Principles}
Four principles guided XWOD's design.
(P1)~\textbf{Real imagery only}: we avoided synthetic augmentation and generative imagery so that reported robustness reflects real-world distributions. (P2)~\textbf{Coverage of climate-amplified hazards}: we expanded beyond the canonical rain/snow/fog triad to include haze/sand/dust, flooding, tornadoes, and wildfire conditions that are becoming more frequent and that lack prior benchmarks. (P3)~\textbf{Geographic diversity}: images were sampled from Asia, North America, Europe, and the United States to limit regional texture bias. (P4)~\textbf{Safety-relevant taxonomy}: we annotated the six traffic categories that dominate vehicle-person and vehicle-vehicle.

\subsection{Data Collection}
Images were sourced from (a) public traffic and weather-event photography archives, and (b) dash-cam video frames donated by driving research partners. Furthermore, the XWOD dataset contains no personal or sensitive information. All images were sourced from public web platforms such as YouTube, adhering to their respective public usage policies. While the dataset includes the Person category as one of its six core traffic classes, these images represent pedestrians in public spaces and are not linked to any personally identifiable information (PII). Sensitive attributes such as gender, age, socioeconomic status, health data or political beliefs were not collected or annotated. (Appendix~\ref{app:datasheet}).

\subsection{Annotation Protocol}
We annotated the six traffic categories using axis-aligned bounding boxes. Following COCO-style protocols, our primary criterion was human recognizability: annotators labeled objects confidently identifiable despite weather degradation, while strictly excluding heavily obscured silhouettes to prevent noisy labels. A comprehensive visual guideline ensured high quality and consistency across all weather domains.

\subsection{Dataset Statistics}
XWOD comprises 10{,}010 images and 42{,}924 labeled instances. Table~\ref{tab:dataset_scale} situates it among prior work. Tables~\ref{tab:weather_images} and \ref{tab:class_distribution} detail per-weather and per-class statistics. Average object count is 4.29 boxes per image; average relative box area of 0.0316, i.e., the benchmark is dominated by small-to-medium objects where weather degradation hurts most.

\begin{table}[t]
\centering
% 1. 使用比 small 更小的專業字體
\footnotesize 
% 2. 縮小欄位間距（預設為 6pt），這是節省橫向空間最有效的方法
\setlength{\tabcolsep}{4pt}
\caption{Scale and provenance of real-image extreme-weather detection datasets. XWOD is the largest by image count and instance count, and the only one covering tornado, flooding, and wildfire.}
\label{tab:dataset_scale}
\begin{tabular}{lrrrcl}
\toprule
Dataset & Year & Images & Instances & \#Weather & Source \\
\midrule
\textbf{XWOD (ours)} & 2026 & \textbf{10{,}010} & \textbf{42{,}924} & \textbf{7} & Asia, N.\ America, Europe, U.S. \\
RTTS \citep{reside2019} & 2018 & 4{,}322 & -- & 1 & China \\
ACDC \citep{acdc2021} & 2021 & 4{,}006 & -- & 4 & Switzerland, central Europe \\
MAS / Foggy Zurich \citep{sakaridis_eccv2018} & 2018 & 3{,}808 & -- & 1 & Zurich, Switzerland \\
WEDGE \citep{wedge2023} & 2023 & 3{,}360 & 16{,}513 & 16 & \textit{Generative (DALL-E)} \\
DAWN \citep{dawn2020} & 2020 & 1{,}000 & 7{,}845 & 4 & Web-sourced \\
\bottomrule
\end{tabular}
\end{table}

\begin{table}[t]
\centering
% 1. 使用比 small 更小的專業字體
\footnotesize 
% 2. 縮小欄位間距（預設為 6pt），這是節省橫向空間最有效的方法
\setlength{\tabcolsep}{4pt}
\caption{Per-weather image and instance counts, with mean boxes per image and mean relative box area. Rare-hazard classes (tornado, flooding, wildfire) contribute 72.7\% of images and 67.8\% of instances---the first time these conditions have been available at scale.}
\label{tab:weather_images}
\begin{tabular}{lrrrr}
\toprule
Weather & Images & Instances & Boxes/Img & Mean Rel.\ Area \\
\midrule
Rain (heavy)   &   665 &  3{,}270 & 4.92 & 0.0244 \\
Snow           & 1{,}203 &  5{,}599 & 4.65 & 0.0213 \\
Fog            &   306 &  1{,}634 & 5.34 & 0.0333 \\
Haze/Sand/Dust &   560 &  3{,}321 & 5.93 & 0.0263 \\
Flooding       & 5{,}151 & 21{,}734 & 4.22 & 0.0368 \\
Tornado        & 1{,}164 &  4{,}962 & 4.26 & 0.0187 \\
Wildfire       &   961 &  2{,}404 & 2.50 & 0.0586 \\
\midrule
\textbf{Total} & \textbf{10{,}010} & \textbf{42{,}924} & 4.29 & 0.0316 \\
\bottomrule
\end{tabular}
\end{table}

\begin{table}[t]
\centering
% 1. 使用比 small 更小的專業字體
\footnotesize 
% 2. 縮小欄位間距（預設為 6pt），這是節省橫向空間最有效的方法
\setlength{\tabcolsep}{4pt}
\caption{Per-weather class distribution (\%). Distributions differ sharply by condition: flooding has the highest pedestrian and VRU share (46.49\% non-vehicle), wildfire contains no motorcycles or bikes, and snow is car-dominated (85.25\%).}
\label{tab:class_distribution}
\begin{tabular}{lrrrrrr}
\toprule
Weather & Car (\%) & Person (\%) & Truck (\%) & Motorcycle (\%) & Bus (\%) & Bike (\%) \\
\midrule
Rain           & 73.15 &  9.17 & 13.18 & 1.80 & 2.20 & 0.49 \\
Snow           & 85.25 &  6.14 &  7.02 & 0.02 & 1.55 & 0.02 \\
Fog            & 68.42 & 15.06 &  8.87 & 0.55 & 6.85 & 0.24 \\
Haze/Sand/Dust & 70.55 & 13.94 &  9.09 & 2.44 & 3.43 & 0.54 \\
Flooding       & 37.56 & 32.37 & 15.17 & 7.23 & 0.77 & 6.89 \\
Tornado        & 64.59 & 27.15 &  7.09 & 0.44 & 0.69 & 0.04 \\
Wildfire       & 48.13 & 37.31 & 12.65 & 0.00 & 1.91 & 0.00 \\
\midrule
\textbf{All}   & 53.94 & 24.77 & 12.17 & 4.06 & 1.47 & 3.59 \\
\bottomrule
\end{tabular}
\end{table}

\subsection{Splits}
We release a fixed train / validation / test partition of $62\% / 15\% / 23\%$ (6{,}206 / 1{,}744 / 2{,}060 images), stratified to preserve the per-weather and per-class distribution within each split. %to within $\pm 3$\,pp. 
Split identifiers are shipped as plain-text index files. For XWOD-Gen (Section~\ref{sec:protocols}), we define an additional set of seven \emph{leave-one-weather-out} splits used for cross-condition generalization.

\section{Evaluation Protocols}
\label{sec:protocols}

XWOD defines two detection-side tracks plus a multimodal weather-understanding probe, each addressing a capability that standard mAP does not measure.

\subsection{XWOD-Det: Standard Detection}

Conventional mAP$_{50}$, mAP$_{50\text{--}95}$, precision and recall at IoU=0.5 in the fixed test split. Serves as the primary leaderboard and is comparable to COCO-style reporting. mAP breakdowns by weather are required for submissions.

\subsection{XWOD-Gen: Leave-One-Weather-Out Generalization}
For each weather $w \in \{1,\dots,7\}$ we train in the other six and test in the holding condition. The reported metric is the \emph{mean} and \emph{worst} per-weather mAP across the seven runs.
\begin{equation}
\text{mAP}^{\text{Gen}} = \tfrac{1}{7}\!\!\sum_{w=1}^{7}\!\! \text{mAP}_{w\mid \bar{w}}, \qquad
\text{mAP}^{\text{Gen}}_{\min} = \min_{w} \text{mAP}_{w\mid \bar{w}}.
\end{equation}

\subsection{XWOD-LLM-WC: Weather Classification of LLMs}
\label{sec:llm-we}
We define an image-only probes for vision-capable LLMs that use only labels already present in the released dataset, no extra annotation.
%(i) \textbf{XWOD-WC}: 
We define a weather classification task, which aims to predict the weather class from $\{$rain, snow, fog, haze/sand/dust, flooding, tornado, wildfire$\}$. Metric: top-1 accuracy, macro-F1, and confusion matrix.
%(ii) \textbf{XWOD-Vis}: predict a visibility bucket $\{L_0, L_1, L_2\}$, where $L_0$ denotes severe occlusion (few or very small visible road users) and $L_2$ denotes a clear scene. Ground-truth visibility is \emph{derived} from the YOLO labels: each image is binned at the global terciles of the total normalized bounding-box area on the chosen split. Metric: accuracy and Spearman $\rho$ on the visibility ordinal. Numerical results will be reported in a forthcoming revision.

\section{Baseline Experiments}
\label{sec:experiments}

\subsection{Setup}
Unless stated otherwise, detectors are trained for 100 epochs at input resolution $640\times 640$ with the Ultralytics default SGD/AdamW schedule. We benchmark the YOLOv8, YOLOv11, and YOLOv26 families at the n/s/m/l/x scales. Experiments are conducted using a combination of a cloud computing environment, dynamically utilizing a pool of professional NVIDIA workstation GPUs (e.g., RTX A4500, and RTX 2000 to 6000 Ada Generation), and a local workstation equipped with an NVIDIA consumer-grade GPU (RTX 4060).

\subsection{Optimization and Generalization}
To address XWOD’s distributional shifts and long-tailed imbalances, we introduce a domain-specific optimization protocol for YOLOv11m.
We use three training choices to improve robustness under extreme weather:
\begin{itemize}[leftmargin=*,topsep=0pt,itemsep=1pt]
    \item \textbf{Robust optimization.} SGD with cosine annealing and $0.1$ label smoothing is used to stabilize training under weather-induced texture degradation.
    \item \textbf{Mitigating long-tailed priors} The classification loss weight is increased to $1.5$, with the learning rate set to $10^{-4}$, to reduce inter-class confusion and improve minority-class learning.
    \item \textbf{Compositional and geometric augmentation.} Rectangular training is disabled, and Copy-Paste ($p=0.2$), MixUp ($p=0.1$), and random rotation within $\pm 10^\circ$ are applied to improve spatial diversity and rare-class exposure.
\end{itemize}

Training efficacy is validated in Figure 4. As shown in Fig. 4(a), it demonstrates superior transferability across synthetic (WEDGE), cross-weather (DAWN), and real-world (RTTS) domains, achieving 63.00\% mAP$_{50}$ on RTTS (vs. 40.37\%), 59.94\% on DAWN (vs. 32.75\%), and 61.12\% on WEDGE (vs. 45.41\%). Per-class analysis in Fig. 4(b) reveals a significant +16.28\% overall gain, with dramatic improvements in rare categories: motorcycle (+25.01\%), bus (+22.92\%), and bike (+17.10\%). These results confirm our optimizations effectively enable learning weather-invariant representations that generalize across unseen domains.

\subsection{YOLO Family Comparison}
Table~\ref{tab:yolo_family} reports detection performance across the three architecture families and five scales. Three observations: (1) the \emph{medium} variant is Pareto-best on XWOD for all three families, contrary to the COCO pattern where larger scales continue to win; we attribute this to the high intra-class texture variance under weather, which regularizes smaller models; (2) YOLOv26l achieves the best overall mAP$_{50\text{--}95}$ of 32.75\%, with YOLOv8m close behind at 32.21\%; (3) YOLOv26m/l distribution focal loss is two orders of magnitude lower than YOLOv8/v11 due to its anchor-free regression redesign.

\begin{table}[t]
\centering
\footnotesize 
\setlength{\tabcolsep}{3pt} % 稍微縮小間距以容納更多欄位
\caption{YOLO family benchmark on XWOD-Det (test split) including loss metrics. Best per architecture in \textbf{bold}; overall best \underline{underlined}. mAP, Precision, and Recall are in \%; Loss values are absolute.}
\label{tab:yolo_family}
\begin{tabular}{ll cccc ccc}
\toprule
\multirow{2}{*}{\textbf{Family}} & \multirow{2}{*}{\textbf{Scale}} & \multicolumn{2}{c}{\textbf{mAP (\%)}} & \multicolumn{2}{c}{\textbf{P / R (\%)}} & \multicolumn{3}{c}{\textbf{Loss}} \\
\cmidrule(lr){3-4} \cmidrule(lr){5-6} \cmidrule(lr){7-9}
& & \textbf{mAP$_{50}$} & \textbf{mAP$_{50\text{--}95}$} & \textbf{Prec.} & \textbf{Rec.} & \textbf{Box} & \textbf{Cls} & \textbf{DFL} \\
\midrule
YOLOv8  & n & 49.24 & 28.74 & 63.59 & 46.79 & 0.9470 & 0.5195 & 0.9642 \\
        & s & 51.53 & 30.60 & 63.42 & 49.23 & 0.7872 & 0.4000 & 0.9099 \\
        & m & \underline{\textbf{54.69}} & \textbf{32.21} & 63.05 & \textbf{53.51} & 0.7538 & 0.3715 &  0.9390 \\
        & l & 52.53 & 31.34 & 68.37 & 47.13 & 0.6877 & 0.3341 & \textbf{0.9292} \\
        & x & 52.70 & 30.97 & 63.08 & 50.28 & 0.6795 & 0.3269 & 0.9274 \\
\midrule
YOLOv11 & n & 46.91 & 27.36 & 65.09 & 43.77 & 0.9345 & 0.5184 & 0.9562 \\
        & s & 48.35 & 28.47 & 60.85 & 47.81 & 0.8129 & 0.4210 & 0.9208 \\
        & m & \textbf{53.84} & \textbf{31.93} & 63.33 & \textbf{53.47} & 0.7642 & 0.3899 & \textbf{0.9253} \\
        & l & 52.06 & 30.86 & 62.33 & 49.75 & 0.7615 & 0.3788 & 0.9511 \\
        & x & 52.64 & 31.41 & 59.97 & 52.39 & 0.7430 & 0.3702 & 0.9556 \\
\midrule
YOLOv26 & n & 46.29 & 26.73 & 57.92 & 45.27 & 1.1891 & 0.6250 & 0.0048 \\
        & s & 51.92 & 30.49 & 63.74 & 50.18 & 1.0087 & 0.4349 & 0.0039 \\
        & m & 53.34 & 32.29 & \underline{\textbf{70.40}} & 48.24 & 0.9066 & 0.3674 & \underline{\textbf{0.0035}} \\
        & l & \textbf{53.95} & \underline{\textbf{32.75}} & 66.24 & \textbf{51.77} & 0.9070 & 0.3627 & \textbf{0.0035} \\
        & x & 53.61 & 32.55 & 67.59 & 51.85 & 0.8717 & 0.3458 & 0.0035 \\
\bottomrule
\end{tabular}
\end{table}

\begin{figure}[h]
\includegraphics[width=0.97\linewidth]{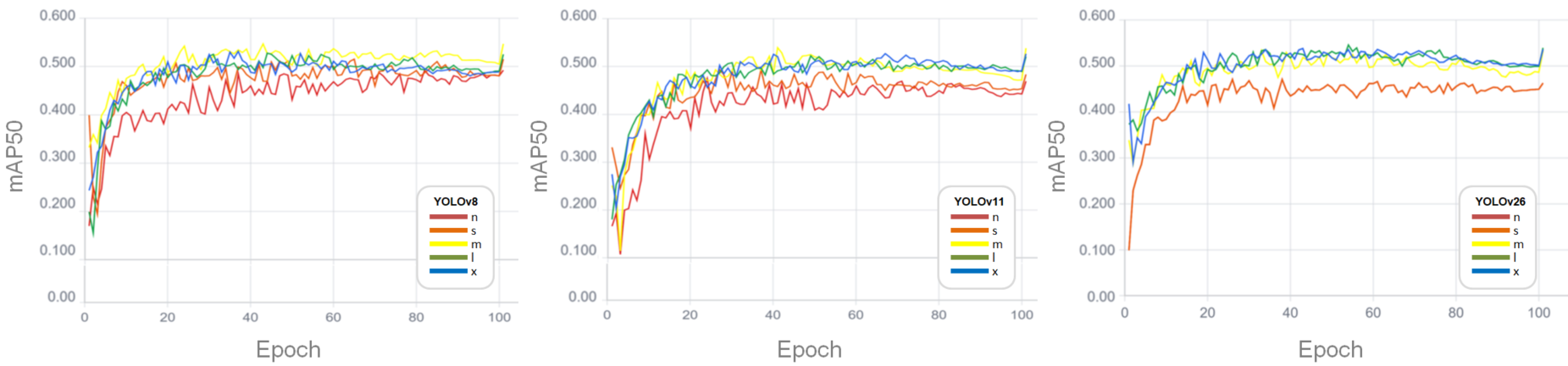}
 \caption{YOLO Family Training Curves.}
\label{fig:scale}
\end{figure}

Scaling and architecture alone are insufficient for extreme weather. Tab. 4 reveal that neither newer models nor larger scales consistently improve XWOD performance. YOLOv8 (49.24–54.69\% mAP$_{50}$) often equals or surpasses YOLOv11 (46.91–53.84\%) and YOLOv26 (46.29–53.95\%), indicating no systematic gains from newer architectures.

Similarly, increasing scale yields marginal returns. Gaps between m, l, and x variants are typically 1–2\% mAP$_{50}$, with medium models frequently outperforming largest counterparts. By breaking the "larger is better" assumption, XWOD reveals critical robustness limits. These marginal gains suggest that dataset diversity, rather than model scale, is the key to detector performance in extreme conditions.

\subsection {Cross-Dataset Comparison}

\begin{figure}[t]
\centering
\includegraphics[width=1\linewidth]{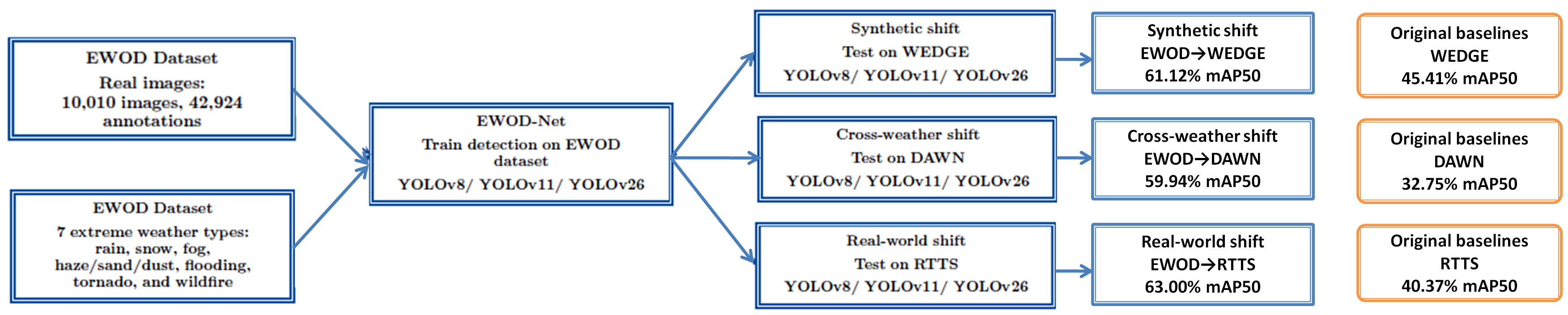}
\caption{Evaluation Protocol of XWOD for weather-aware domain adaptation.}
\label{fig:domain-adapt}
\end{figure}

Table~\ref{tab:ewod_cross_domain} compares XWOD with established extreme-weather benchmarks. Figure \ref{fig:domain-adapt} shows the cross-dataset evaluation protocol. In contrast to benchmarks focused on specific conditions (e.g., RTTS haze or WEDGE synthetic imagery), XWOD captures a broader real-world spectrum, including climate-amplified hazards such as flooding, tornadoes, and wildfires. 

The cross-dataset results indicate that XWOD can serve as an effective source domain for weather-robust traffic-object detection. A detector trained only on XWOD achieves 61.12\% mAP$_{50}$ on WEDGE, 59.94\% on DAWN, and 63.00\% on RTTS, compared with the corresponding published baselines of 45.41\%, 32.75\%, and 40.37\%, respectively. These results correspond to relative improvements of 34.6\%, 83.0\%, and 56.1\%. Since the published baselines differ in model architecture, training data, and evaluation protocol, these comparisons should be interpreted as evidence of XWOD's transfer utility rather than as a controlled detector-level comparison.

\begin{figure}[h]
\includegraphics[width=1\linewidth]{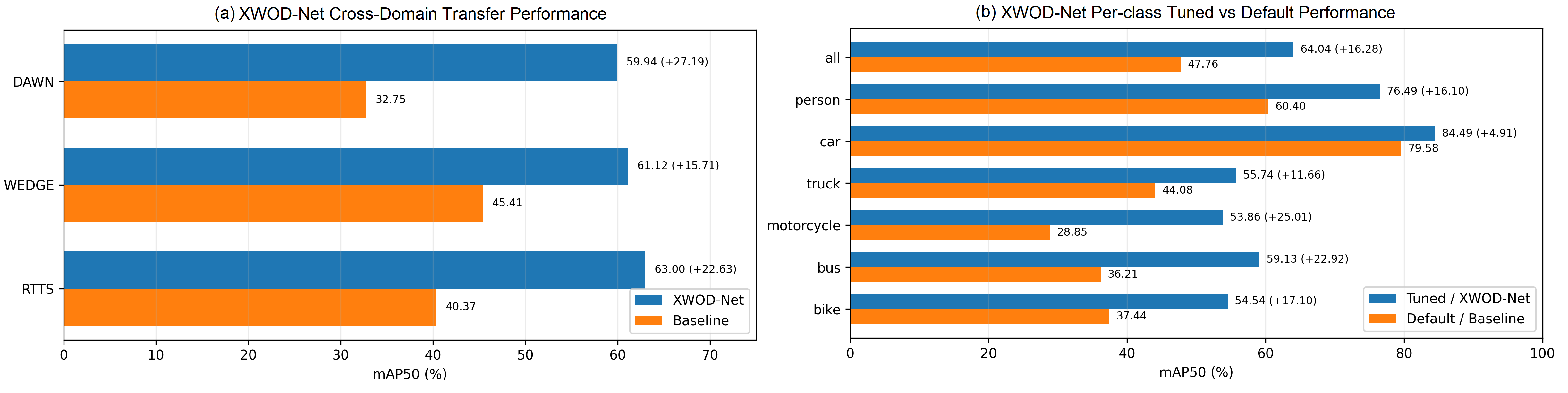}
\caption{(a) YOLOv8m demonstrates strong generalization across synthetic (WEDGE), cross-weather (DAWN), and real-world (RTTS) domains, significantly outperforming native baselines. (b) YOLOv11m optimization protocol yields a +16.28\% overall mAP$_{50}$ gain, with substantial improvements observed in rare categories such as motorcycle (+25.01\%) and bus (+22.92\%).}
\label{fig:cross}
\end{figure}

\begin{table}[ht]
\centering
% 1. 使用比 small 更小的專業字體
\footnotesize 
% 2. 縮小欄位間距（預設為 6pt），這是節省橫向空間最有效的方法
\setlength{\tabcolsep}{6pt}
\caption{Cross-domain evaluation against established baselines. Our XWOD models are trained on the XWOD dataset and tested zero-shot on unseen domains.}
\label{tab:ewod_cross_domain}
\begin{tabular}{lcccccccc}
\toprule
\textbf{Category} & \textbf{Setting / Source} & \textbf{Model / Baseline} & \textbf{mAP\textsubscript{50} (\%)}  & \textbf{Precision (\%)} & \textbf{Recall (\%)} \\
\midrule

\multicolumn{6}{c}{\textbf{In-domain (XWOD-Dataset)}} \\
XWOD-Net & XWOD $\rightarrow$ XWOD & \textbf{YOLOv8m} & \textbf{54.69} & 63.05 & 53.51 \\
& & YOLOv11m & 53.87 & 63.33 & 53.47 \\
& & YOLOv26m & 53.34 & 70.40 & 48.24 \\
& & YOLOv8l & 52.53 & 68.37 & 47.13 \\
& & YOLOv11l & 52.06 & 62.33 & 49.75 \\
& & YOLOv26l & 53.95 & 66.24 & 51.77 \\

\midrule
\multicolumn{6}{c}{\textbf{Synthetic Domain Shift}} \\
\textbf{Baseline} & \textbf{WEDGE Original} & \textbf{Faster R-CNN} & \textbf{45.41} & - & - \\
XWOD-Net & XWOD $\rightarrow$ WEDGE & \textbf{YOLOv8m} & \textbf{61.12} & 66.23 & 55.98 \\
& & YOLOv11m & 55.02 & 64.33 & 53.68 \\
& & YOLOv26m & 53.94 & 71.83 & 49.00 \\
& & YOLOv8l & 56.22 & 64.08 & 50.54 \\
& & YOLOv11l & 55.29 & 64.16 & 53.95 \\
& & YOLOv26l & 54.90 & 68.76 & 51.74 \\

\midrule
\multicolumn{6}{c}{\textbf{Cross-Weather Generalization}} \\
\textbf{Baseline} & \textbf{DAWN Original} & \textbf{Ensemble Det.} & \textbf{32.75} & - & - \\
XWOD-Net & XWOD $\rightarrow$ DAWN & \textbf{YOLOv8m} & \textbf{59.94} & 65.73 & 52.74 \\
& & YOLOv11m & 55.17 & 61.81 & 52.18 \\
& & YOLOv26m & 54.81 & 66.58 & 52.62 \\
& & YOLOv8l & 55.28 & 66.37 & 53.95 \\
& & YOLOv11l & 50.56 & 61.21 & 48.78 \\
& & YOLOv26l & 56.99 & 72.08 & 52.08 \\

\midrule
\multicolumn{6}{c}{\textbf{Real-world Domain Adaptation}} \\
\textbf{Baseline} & \textbf{RTTS Original} & \textbf{Standard Det.} & \textbf{40.37} & - & - \\
XWOD-Net & XWOD $\rightarrow$ RTTS & \textbf{YOLOv8m} & \textbf{63.00} & 69.70 & 56.56 \\
& & YOLOv11m & 54.57 & 69.65 & 50.67 \\
& & YOLOv26m & 54.95 & 69.70 & 56.56 \\
& & YOLOv8l & 60.31 & 64.53 & 55.09 \\
& & YOLOv11l & 57.29 & 66.22 & 53.70 \\
& & YOLOv26l & 56.74 & 67.95 & 53.07 \\

\bottomrule
\end{tabular}
\end{table}

Table~\ref{tab:cross_dataset} further situates XWOD among related adverse-weather datasets. XWOD obtains 54.69\% mAP$_{50}$ on its in-domain test set while covering seven weather conditions, including three rare climate-amplified hazards that are absent from prior real-image detection benchmarks. We therefore report XWOD primarily as a detection benchmark and treat segmentation-oriented datasets such as ACDC and MAS as complementary references rather than directly comparable quantitative baselines. Overall, the results suggest that real-world weather diversity improves cross-domain transfer and exposes failure modes that are not captured by narrower or synthetic benchmarks.

\begin{table}[t]
\centering
% 1. 使用比 small 更小的專業字體
\footnotesize 
% 2. 縮小欄位間距（預設為 6pt），這是節省橫向空間最有效的方法
\setlength{\tabcolsep}{4pt}
\caption{Cross-dataset comparison of best published detection results. ``---'' indicates the paper does not report that metric.}
\label{tab:cross_dataset}
\begin{tabular}{llccc}
\toprule
Dataset & Best Model & mAP (\%) (All) & Best Per-Weather (\%) & Notes \\
\midrule
\textbf{XWOD (ours)}   & YOLOv8m          & \textbf{54.69} & 62.93 (Tornado) & 7 weather \\
                  & YOLOv11m         & 53.84 & 68.97 (Tornado) & 7 weather \\
                  & YOLOv26l         & 53.95 & 66.01 (Tornado) & 7 weather \\
RTTS \citep{reside2019}  & --              & 40.37 & --- & 1 weather \\
%ACDC \citep{acdc2021}  & ResNet-50           & 53.40 (mIoU) & 57.60 (Rain) & 3 weather \\
DAWN \citep{dawn2020}  & Ensemble         & 32.75 & --- & 4 weather \\
%MAS \citep{sakaridis_eccv2018} & --      & 34.30 (mIoU) & --- & 1 weather \\
WEDGE \citep{wedge2023}& Faster R-CNN     & 22.78 & --- & 16 labels, generative \\
                  & Fine-tuning      & 45.41 & --- & \\
\bottomrule
\end{tabular}
\end{table}

\subsection{Per-Weather Breakdown}
Table~\ref{tab:per_weather} reports YOLOv11m performance by weather, revealing a $51.12\,$pp gap between the best (tornado, 68.97\% mAP$_{50}$) and worst (wildfire, 17.85\%) conditions. Wildfire and fog dominate the error budget: wildfire degrades recall to 15.40\%, the model finds only 1 in 6 objects, and fog halves recall relative to rain. These are the rare, safety-critical cases existing benchmarks never surfaced.

\begin{table}[t]
\centering
% 1. 使用比 small 更小的專業字體
\footnotesize 
% 2. 縮小欄位間距（預設為 6pt），這是節省橫向空間最有效的方法
\setlength{\tabcolsep}{4pt}
\caption{Per-weather performance of YOLOv11m on XWOD-Det. Wildfire and fog are the clear failure modes. ``Loss'' columns from the final training epoch on each condition.}
\label{tab:per_weather}
\begin{tabular}{lcccccccc}
\toprule
Weather & mAP$_{50}$ (\%)  & mAP$_{50\text{--}95}$ (\%) & Prec. (\%) & Rec. (\%) & Box (loss) & Cls (loss) & DFL (loss) \\
\midrule
Rain           & 60.49 & 37.70 & 61.74 & 61.87 & 0.699 & 0.406 & 0.938 \\
Snow           & 38.46 & 25.93 & 69.79 & 32.19 & 0.534 & 0.314 & 0.810 \\
Fog            & 23.45 & 11.97 & 66.08 & 21.92 & 0.976 & 0.564 & 1.168 \\
Haze/Sand/Dust & 29.86 & 9.95 & 38.83 & 35.10 & 0.650 & 0.390 & 0.888 \\
Flooding       & 42.86 & 26.36 & 53.15 & 43.73 & 0.740 & 0.375 & 0.919 \\
Tornado        & \textbf{68.97} & \textbf{43.47} & \textbf{75.38} & \textbf{57.70} & 0.771 & 0.415 & 0.905 \\
Wildfire       & 17.85 & 12.18 & 60.57 & 15.42 & 0.409 & 0.279 & 0.845 \\
\bottomrule
\end{tabular}
\end{table}

\subsection{Weather Classification by LLMs}

We also explore the image understanding ability of modern LLMs, and design a simple task that requests LLMs to classify the weather in our images. The task goal is that given a single street-level image, the LLM model must predict the dominant extreme weather condition as one of seven canonical labels: \emph{rain}, \emph{snow}, \emph{fog}, \emph{haze/sand/dust}, \emph{flooding}, \emph{tornado},\emph{wildfire}. Ground truth is read directly from the XWOD filename prefix (\texttt{heavy}\,$\to$\,rain, \texttt{dust}\,$\to$\,haze/sand/dust,etc.) and requires no additional annotation. The evaluation prompt is defined below. All vision-language models receive an identical zero-shot, single-turn prompt (no in-context examples, no chain of thought):                                                                                      
  \begin{quote}\small\itshape
  You are an expert in autonomous-driving scene understanding. Look at the image and classify the primary extreme weather condition as ONE of:                             rain $\mid$ snow $\mid$ fog $\mid$ haze/sand/dust $\mid$ flooding                    $\mid$ tornado $\mid$ wildfire. Reply with only the label, lowercase, no punctuation.         
  \end{quote}                                                                                    
  The free-text reply is canonicalized via exact match, then a small
  keyword fallback (e.g.\ \emph{foggy}\,$\to$\,fog, \emph{flood}\,$\to$\,flooding, \emph{fire}\,$\to$\,wildfire) before scoring; unmatched outputs are kept as \texttt{UNKNOWN}.
  
  Evaluation runs on the XWOD test split. To remove the long-tail imbalance from the metric, we sample a class-balanced probe of $100$ images per class ($7\times 100 = 700$ images total) using a fixed seed; images are re-shuffled before querying to avoid order effects. The whole probe is served to every model identically; per-image inputs are downscaled so the longest side is at most $1024$\,px and re-encoded as JPEG quality  90 to bound API token cost. Due to the limit of time and budget, we only select 4 latest LLM models for evaluation. The test results are shown in Table \ref{tab:llm_wc}. All mainstream models have achieved high accuracy on the task.  The top 1 and top 2 models are from Google Gemini family, followed by Claude Opus 4.7 and ChatGPT 5.5.

\begin{table}[h]
\centering
\caption{Weather classification performance of commercial LLMs.}
\begin{tabular}{llcc}
\toprule
Provider & Model Name      & Accuracy & Macro F1 \\
\midrule
Google      & gemini-3.1-pro-preview        & 0.7571       & 0.7585        \\
Google      & gemini-3.1-flash-lite-preview & 0.7500       & 0.7500        \\
Anthropic   & claude-opus-4-7               & 0.7471       & 0.7446        \\
OpenAI      & gpt-5.5                       & 0.7186       & 0.7221       \\
\bottomrule
\end{tabular}
\label{tab:llm_wc}
\end{table}

\section{Discussion}

\paragraph{Wildfire and fog are open problems.}
XWOD detectors that score above 54.69\% mAP$_{50}$ overall fall below 17.85\% on wildfire and 23.45\% on fog. The two conditions share the \emph{contrast-suppression} failure mode: both collapse object–background color gradients and reduce mid-frequency texture, rendering anchor-based heads brittle. This motivates restoration-coupled detectors (IA-YOLO, TogetherNet, DENet), transformer backbones (RT-DETR, D-FINE), and open-vocabulary detectors (YOLO-World, Grounding DINO) as natural next candidates on XWOD-Det and XWOD-Gen.

\paragraph{Class prior versus weather prior.}
Flooding contains 6.90\% bike instances vs.\ 0.02--0.55\% in every other condition; wildfire contains \emph{zero} motorcycle or bike instances. These asymmetric priors make leave-one-weather-out generalization (XWOD-Gen) a stringent test: models must avoid overfitting the joint (class, weather) prior while still exploiting it where correct.

\paragraph{Real versus synthetic.}
XWOD’s real imagery enables direct comparison with WEDGE’s VLM-generated scenes. Cross-evaluation could quantify the "real-data premium." We anticipate synthetic-trained detectors will transfer poorly to XWOD’s flooding and wildfire splits, where generative models produce plausible but statistically mismatched scenes.

\section{Limitations}

XWOD has four main limitations. First, it is a 2D camera-based object detection benchmark and does not include LiDAR or radar signals commonly used in production autonomous-driving stacks; multimodal weather datasets such as STF \citep{stf2020}, CADC \citep{cadc2021} and Ithaca365 \citep{ithaca365} are therefore complementary. Second, XWOD provides annotations for six traffic categories rather than the 19 classes used in some driving benchmarks, and is not designed for semantic segmentation. Third, although XWOD covers four continents, the weather distribution is imbalanced: flooding accounts for 51.5\% of the images due to the greater availability of public flood imagery. We therefore report per-weather results to avoid masking this imbalance with aggregate metrics. Finally, tornado samples are limited to post-event or near-event scenes in which the tornado is visible in the frame. Pre-tornado conditions are not included.

\section{Conclusion}
We introduce XWOD, the largest real-image traffic object-detection benchmark for extreme weather to date. XWOD covers seven weather conditions and is the first benchmark to include the climate-amplified hazards of flooding, tornadoes, and wildfires. In addition to standard mAP evaluation, we define two detection protocols, including leave-one-weather-out generalization (XWOD-Gen), and a multimodal weather-understanding probe (XWOD-LLM-WC). Our main empirical result is that XWOD provides a strong source domain for transfer learning: detectors trained \emph{only} on XWOD outperform the published baselines of each target dataset under zero-shot evaluation on RTTS (63.00\% vs.\ 40.37\%), DAWN (59.94\% vs.\ 32.75\%), and WEDGE (61.12\% vs.\ 45.41\%). In-domain evaluation further reveals a 51.12-percentage-point gap across weather conditions, with wildfire and fog remaining the most challenging cases. XWOD is released with a datasheet, evaluation protocols, baseline weights, and reproducible code.
\bibliographystyle{plainnat}
\bibliography{ewod}

\end{document}